\documentclass{svproc}
\usepackage{url}
\usepackage{graphicx}
\usepackage[square,sort,comma,numbers]{natbib}
\usepackage{amsmath}

\begin{document}
\mainmatter              
\title{A comparison of LSTM and GRU networks
for learning symbolic sequences\thanks{R.\,C.'s work has been funded by the UK's Alan Turing Institute. S.\,G.\ has been supported by a Fellowship of the Alan Turing Institute, EPSRC grant EP/N510129/1.}}
\titlerunning{A comparison of LSTM and GRU networks
for learning symbolic sequences}  
%
%
%

%
\author{$\text{Roberto Cahuantzi}^{[0000-0002-0212-6825]}$ \and $\text{Xinye Chen}^{[0000-0003-1778-393X]}$ \and $\text{Stefan G\"{u}ttel}^{[0000-0003-1494-4478]}$}
\authorrunning{Roberto Cahuantzi et al.} 
\tocauthor{Roberto Cahuantzi, Xinye Chen, and Stefan G\"{u}ttel}
\institute{The University of Manchester, Department of Mathematics,\\ Manchester, M13 9PL, United Kingdom}

\maketitle              

\begin{abstract}
We explore the architecture of recurrent neural networks (RNNs) by studying the complexity of string sequences it is able to memorize. Symbolic sequences of different complexity are generated to simulate RNN training and study parameter configurations with a view to the network's capability of learning and inference. We compare Long Short-Term Memory (LSTM) networks and gated recurrent units (GRUs). We find that an increase in RNN depth does not necessarily result in better memorization capability when the training time is constrained. Our results also indicate that the learning rate and the number of units per layer are among the most important hyper-parameters to be tuned. Generally, GRUs outperform LSTM networks on low-complexity sequences while on high-complexity sequences LSTMs perform better. 
\end{abstract}

\section{Introduction}
The recurrent neural network (RNN) is an extremely expressive sequential model to learn sequence data and plays an important role in sequence-to-sequence learning such as image captioning \citep{mao2014deep, 7410872},  speech modeling \citep{45168}, symbolic reasoning tasks \citep{NIPS2014_08419be8, Lample2020Deep, Kim_Nam_Kim_Jung_2021}, and time series prediction \citep{10.1109/ICASSP.2017.7953302, Elsworth2020lstm}.  Reliable and computationally efficient methods to forecast trends and mining the patterns in sequence data are very desirable; Recent sequential models achieve significant success in temporal sequence forecasting; see e.g. \cite{SALINAS20201181} which introduces a probabilistic forecasting methodology based on an autoregressive recurrent neural network model. An interpretable deep learning time series prediction framework is proposed in \cite{Oreshkin2020NBEATSNB}. A lot of efforts have also gone into studying the architecture of the sequential models; see e.g. \cite{pmlr-v37-jozefowicz15} which gives an empirical exploration on RNN by conducting a thorough architecture search over different RNN architectures. \cite{10.5555/3172077.3172204} compares a sophisticated hybrid neural network model to simpler network models and more traditional statistical methods (such as hidden Markov models) for trend prediction, with the hybrid model achieving the best results. Another hybrid forecasting method that combines RNNs and exponential smoothing is discussed in~\cite{smyl2020hybrid}. Comparisons of LSTM and GRU networks on numerical time series data tasks can be found in \cite{yamak2019comparison}. 

Despite these significant advances in the sequential models, there is also growing literature suggesting that data pre-processing is just as important to the performance as model architecture. In this realm, \cite{rabanser2020} shows that discretization of data can improve the forecasting performance of neural network models. Another critical aspect is the metrics to evaluate the performance of models in forecasting or other inference tasks. In these settings, the Euclidean distance metric and its variants, such as the mean squared error, are often used in this context. However, these metrics can be sensitive to noise in the data, an effect that becomes even more pronounced with time series of high dimensionality. Hence, \cite{Wang2005, Elsworth2020lstm} argue that symbolic time series representations, which naturally offer dimensionality reduction and smoothing, are useful tools to allow for the use of discrete (i.e.\ symbolic) modeling.

Here we give empirical insights into the connections between the hyper-parameters of popular RNNs and the complexity of the string sequences to be learned (and forecasted). This study is partly inspired by \cite{Greff2017} who evaluate the performance of many variants of LSTM cells via extensive tests with three benchmark problems, all rather different from our string learning task. Among our main findings are that: (1) the learning rate is one of the most influential parameters when training RNNs to memorize sequences (with values near $10^{-2}$ found to be the best in our setup in terms of training time and forecast accuracy); (2) for the tasks considered here it is often sufficient to use just common RNNs with a single layer and a moderate number of units (such as around 100 units); (3) GRUs outperform LSTM networks on low complexity sequences while on high complexity sequences the order is reversed. The Python code used to perform our experiments is publicly available\footnote{\url{https://github.com/robcah/RNNExploration4SymbolicTS}}. To facilitate the community to study machine learning models on symbolic sequences, based on this research, related methods in this paper have been included in a Python library \texttt{slearn} \citep{Cahuantzi_slearn_2021} that enables producing synthetic symbolic sequences of user-specific complexity and comparative study of models. 

Note that another common approach to using deep learning for regression is global forecasting models (GFMs) which are employed on a large scale of temporal data; see e.g. \cite{Bandara2020, Monteromanso2021}. While this method is appealing due to the improved generalizability of the resulting models, which enables reduced proneness to overfit and potentially lower overall training time. However, the model complexity is significantly higher and the selection of hyper-parameters is even more involved. Therefore, it is hard to understand the relationship between model architecture and input complexity. Here we take a different, simpler, approach by training one RNN model at a time to learn the symbolic sequence of various string complexity. This will provide more direct insight into the learning capability of a single RNN dependent on the complexity of the symbolic sequences it is meant to learn. Since GFM is expected to be at least as complex as the model required to learn, we believe that our study also sheds light on some parameter choices for GFMs.

\section{Methodology}
Our approach to quantifying the learning capabilities of RNNs is to generate symbolic sequences of different string complexities (with complexity measured in terms of the compressibility of the string), to train RNNs on a part of that string until a predefined stopping criterion is reached, and then to quantify the accuracy of the forecast of the following string characters in an appropriate text similarity metric. Below we provide details for each of these steps.

\subsection{String generation and LZW complexity}
As training and test data for this study, we produce a collection of strings with quantifiable complexities. These strings are here-forth referred to as \emph{seed strings}. A Python library was written to generate these seed strings, allowing the user to choose the target complexity and the number of distinct symbols to be used. 

One way to quantify complexity is due to Kolmogorov~\cite{Kolmogorov1963}: the length of the shortest possible description of the string in some fixed universal language without losing information. For example, a string with a thousand characters simply repeating \verb|"ab"| can be described succinctly as \verb|500*"ab"|, while a string of the same length with its characters chosen at random does not have a compressed representation; therefore the latter string would be considered more complex. 

A more practical approach to estimate complexity uses lossless compression methods \citep{Kaspar1987,Zenil2020}. The Lempel–Ziv–Welch (LZW) compression \cite{Welch1984} is widely recognized as an approximation to Kolmogorov complexity. The LZW algorithm serves as the basis of our complexity metric as it is very easy to implement and can be adapted to generate strings of a target compression rate. The LZW algorithm creates a dictionary of substrings and an array of dictionary keys from which the original string can be fully recovered. 
We define the \emph{LZW complexity} of a seed string as the length of its associated LZW array, an upper bound on the Kolmogorov complexity.
\begin{figure}[h]
	\vspace*{-5mm}
	\centering
	\includegraphics[height=0.125\textheight]{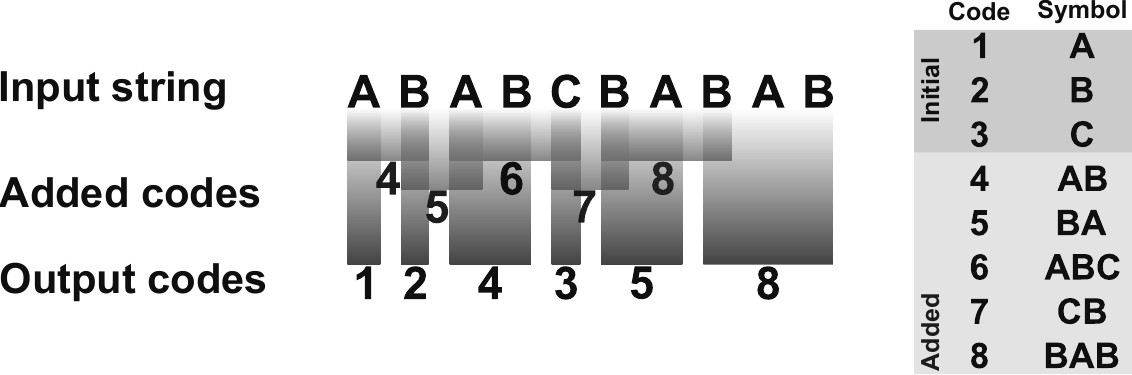}
	\caption[Illustration of LZW compression]{An illustration of LZW compression. Assume that we have an alphabet of three symbols ($A, B, C$) and the seed string \texttt{"ABABCBABAB"}. As the LZW algorithm traverses the seed string from left to right, a dictionary of substrings is built (table on the right). If a combination of characters already contained in the dictionary is found, the related index substitutes the matching substring. In this case, the resulting array is $[1,2,4,3,5,8]$ corresponding to an LZW complexity of 6.}
	\label{fig:LZW00}
\end{figure}


\subsection{Training, test, and validation data}
The data used to train and evaluate the RNN models is obtained by repeating each seed string until a string $s$ of predefined minimal string length is reached. The trailing $n$ characters of $s$ are split off to form a validation string~$v$. The remaining leading characters are traversed with a sliding window of $n$~characters to produce input and output arrays $X$ and $y$, respectively, for the training and testing. Here, the input array $X$ is of dimension $m\times n\times p$, where $m$ stands for the number of input sequences ($m=|s|-2 n$ where $s$ denotes the length of $s$), $n$ is the length of each input sequence, and $p$ is the dimension of the binary vectors used for the one-hot encoding of each of the distinct characters. 
The output array $y$ contains the next symbol following each string sequence encoded in $X$ and is of dimension $m\times p$. The pair $(X,y)$ is split 95\% vs 5\% to produce the training and test data, respectively. (This rather low fraction of test data is justified as there occur repeated pairs $(X,y)$ in the data due to the repetitions in the string $s$.) The test data is used to compute the RNN accuracy and loss function values. Finally, the one-hot encoding of the validation string $v$ results in an array of dimension $n\times p$. The trained RNN model is then used to forecast the validation string $v$, and a text similarity measure quantifies the forecast accuracy.


To exemplify this we can imagine a seed string \verb|"abc"| with $p=3$ distinct characters, which will be repeated to reach a string $s$ of at least 100 characters length. In this case, $s=$\verb|"abcabcabc...abc"| is of length 102 characters. The trailing $n=10$ characters are split off for the validation, resulting in $v=$\verb|"cabcabcabc"|. The remaining 92 leading characters of $s$ are then traversed with a sliding window of width $n=10$ to form the input-output data pairs $(X,y)$ as follows:
\begin{verbatim}
	(abcabcabca,b) (bcabcabcab,c) (cabcabcabc,a) ... (abcabcabca,b)
\end{verbatim} 
The one-hot encoding $a=[1,0,0]$, $b=[0,1,0]$, $c=[0,0,1]$ results in the final arrays used for the training and testing.

\subsection{Recurrent Neural Networks}

We consider two types of RNN architecture, i.e., Long Short-Term Memory (LSTM) cells~\citep{Hochreiter1997} and Gated Recurrent Units (GRUs) \citep{cho2014}, respectively. Different versions of these units exist in the literature, so we briefly summarize the ones used here. 

\begin{figure}[h]
	\centering
	\includegraphics[width=0.2\textwidth]{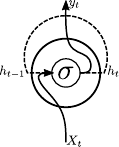}
	\includegraphics[width=0.6\textwidth]{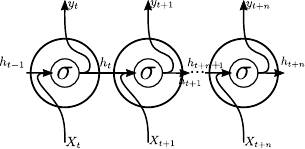}
	\caption{A simple RNN cell on a single time-step (left) and the unfolded interpretation of the same RNN (right).}
	\label{fig:RNN}
\end{figure}

A standard LSTM cell includes three \emph{gates}: the forget gate $f_t$ which determines how much of the previous data to forget; the input gate $i_t$ which evaluates the information to be written into the cell memory; and the output gate $o_t$ which decides how to calculate the output from the current information:
\begin{equation} \label{eq:LSTM01}
	\begin{split}
		i_t &=\sigma(W_i \ X_t+R_i \ h_{t-1}+b_i)\\
		f_t &=\sigma(W_f \ X_t+R_f \ h_{t-1}+b_f)\\
		o_t &=\sigma(W_o \ X_t+R_o \ h_{t-1}+b_o).
	\end{split}
\end{equation}
Here, the $W, R$, and $b$ variables represent the matrices and vectors of trainable parameters. 
The LSTM unit is defined by
\begin{equation} \label{eq:LSTM00}
	\begin{split}
		\dot{C_t} &=\tanh(W_c \ X_t+R_c \ h_{t-1}+b_c)\\
		C_t &=f_t \odot C_{t-1}+i_t \odot \dot{C_t}\\
		h_t &=o_t\odot \tanh(C_t)\\
		y_t &=\sigma(W_y \ h_t+b_y).
	\end{split}
\end{equation}
In words, the candidate cell state $\dot{C_t}$ is calculated using the input data $X_t$ and the previous hidden state $h_{t-1}$. The cell memory or current cell state $C_t$ is calculated using the forget gate $f_t$, the previous cell state $C_{t-1}$, the input gate $i_t$ and the candidate cell state $\dot{C_t}$. The Hadamard product $\odot$ is simply the element-wise product of the involved matrices. The output $y_t$ is calculated by applying the corresponding weights ($W_y$ and $b_y$) to the hidden state $h_t$.

\begin{figure}[h]
	\centering
	\includegraphics[width=0.47\textwidth]{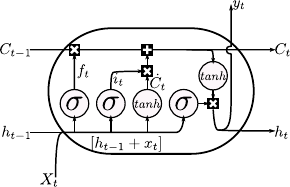}
	\includegraphics[width=0.47\textwidth]{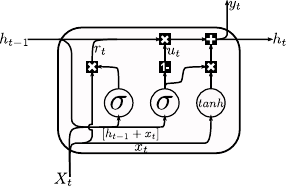}
	\caption{General structure of LSTM (left) and GRU (right) units.}
	\label{fig:LSTM_GRU}
\end{figure}

GRUs are similar to LSTMs but use fewer parameters and only two gates: the update ($u_t$) and reset ($r_t$) gates. The gate $u_t$ tunes the update speed of the hidden state while the gate $r_t$ decides how much of the past information to forget by resetting parts of the memory. The GRU unit is defined by the below set of equations. In them $\dot{h_t}$ stands for the candidate hidden state. 
\begin{equation} \label{eq:GRU00}
	\begin{split}
		u_t &=\sigma(W_u \ x_t+R_u \ h_{t-1}+b_u)\\
		r_t &=\sigma(W_r \ x_t+R_r \ h_{t-1}+b_r)\\
		\dot{h_t} &=\tanh(W_h \ x_t+ (r_t \odot h_{t-1})R_h+b_h)\\
		h_t &=(1-u_t)\odot h_{t-1}+u_t \odot \dot{h_t}\\
		y_t &=\sigma(W_y \ h_t+b_y) 
	\end{split}
\end{equation}

\figurename~\ref{fig:RNN} and \figurename~\ref{fig:LSTM_GRU} illustrate the general RNN architecture and its variants LSTM and GRU.

\subsection{Text similarity metrics}

Our experimental tests are performed on string sequence rather than numerical data (i.e., associated with quantitative values), so the well-justified text similarity metrics are necessary for the reasonable assessment and convincing conclusions. Here we briefly discuss the metric we used in this paper for string prediction accuracy. Due to the non-Euclidean nature of symbolic representations, the accuracy of the forecast is best quantified via text edit metrics such as the Damerau--Levenshtein (DL) and Jaro--Winkler (JW) distance. The DL distance counts the number of edit steps required to transform a string into another \citep{Boystov2011}. The JW distance is a more elaborate metric that is less sensitive to string insertions and changes in character positions; see \cite{Winkler2006}.

The following explains briefly the text distance algorithms used on this project, to give an intuitive understanding of these metrics. According to \cite{Boystov2011} the Damerau-Levenshtein (DL) text distance can be formalised with the algorithm \ref{eq:damerau00}. 
\begin{equation} \label{eq:damerau00}
	\begin{split}
		dl_{a,b}(i,j) = \min \begin{cases}
			0,  & \text{if} \ i=j= 0,\\ 
			dl_{a,b}(i-1,j) + 1 & \text{if} \ i>0 \ \text{(deletion)},\\
			dl_{a,b}(i,j-1) + 1 & \text{if} \ j>0 \ \text{(insertion)},\\
			dl_{a,b}(i-1,j-1) + 1_{(a_i\ne b_j)} & \text{if} \ i>0 \ \text{and} \ j>0 \ \text{(substitution)},\\
			dl_{a,b}(i-2,j-2) + 1 & \text{if} \ i>1 \ \text{and} \ j>1 \ \text{and} \ a_{i} = b_{j-1} \\& \text{and} \ a_{i-1} = b_{j} \ \text{(transposition)}.
		\end{cases}
	\end{split}
\end{equation}

Here, $dl_{a,b}(i,j)$ means distance between the first $i$ characters of $a$ and first $j$ characters of $b$. The symbols $a_i$ and $b_j$ stand for the character of the strings in positions $i$ and $j$ respectively. The expression $1_{(a_i\ne b_j)}$ is the conditional value 0 if $a_i = b_j$ but 1 otherwise. Jaro-Winkler distance (JW), from \cite{Winkler2006}, is symbolised as $d_{jw}$, and based on Jaro similarity ($sim_{j}$), the latter being defined with equation \ref{eq:jaro00}.

\begin{equation} \label{eq:jaro00}
	sim_{j} = \frac{1}{3} \Big( \frac{m}{|s_1|} + \frac{m}{|s_2|} + \frac{m-t}{m} \Big)
\end{equation}
\begin{equation} \label{eq:matchingtol01}
	d_{max} < \Big \lfloor \frac{max(|s_1|, |s_2|)}{2} \Big \rfloor - 1
\end{equation}

Where $|s_i|$ is the length of string $s_i$, while $t$ stands for the number of transpositions (the matching characters in different sequence order divided by $2$), and $m$ is the number of matching characters only if the distance ($d_{max}$) obeys equation \ref{eq:matchingtol01}. Equation \ref{eq:jarowinkler00}, defines JW distance ($d_{jw}$), $\ell$ stands for the length of common prefix at the start of the string up to four characters, $p$ is a scaling factor for how much the score is adjusted upwards for common prefixes, it should not exceed 0.25 and the standard value is 0.1.
\begin{equation} \label{eq:jarowinkler00}
	d_{jw} = 1 - [sim_j + \ell \ p (1-sim_j)]
\end{equation}

\begin{figure}[h]
	\centering
	\includegraphics[width=1.0\textwidth]{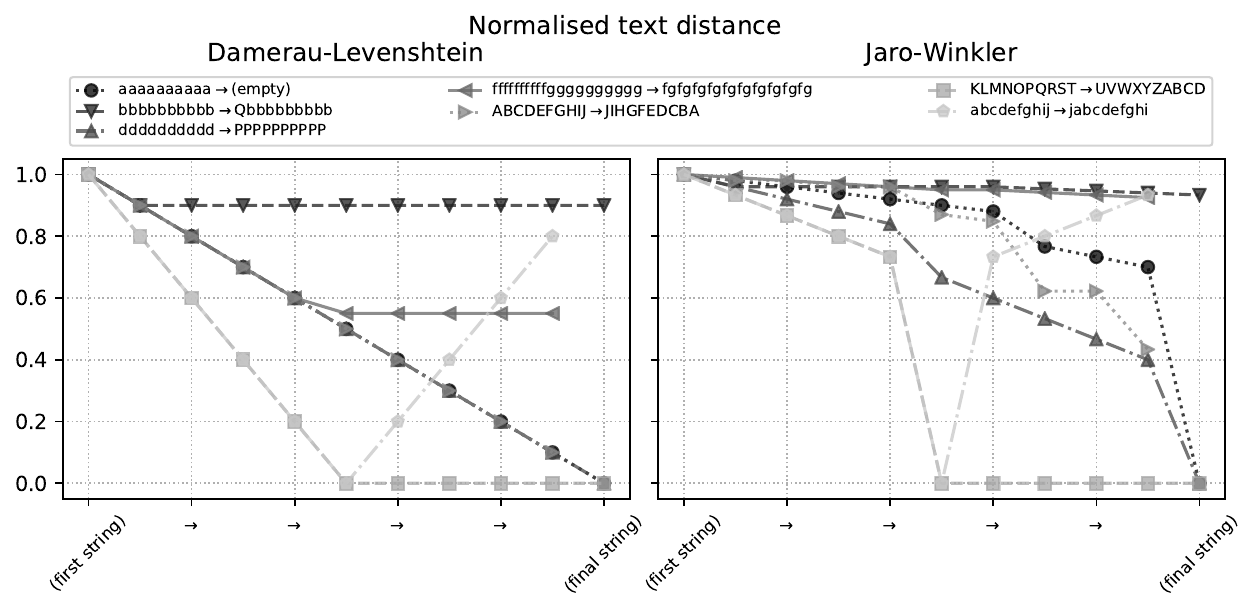}
	\caption[Text distance illustration]{Text distance illustration: the x-axis remain empty because it accounts for the progression of the first string towards the last one of the sets seen in the upper legend. }
	\label{fig:textdistance00}
\end{figure}

For clearness these values are inversely normalized to the highest distance between strings, meaning a value of $0$ for completely different strings and a value of $1$ for completely matching ones. Figure \ref{fig:textdistance00} summarises these values between several progressions of different strings. Meaning, the normalized text distance between the initial string to the progression to the final one. For the first string $aaaaaaaaaa\rightarrow (empty)$, the transitional values ($x$ axis) correspond to strings with one less $a$ character by step until reaching an empty string. For most of the progressions, the first string was gradually overtaken by the final one, with exception of the fourth string in which two symbols ($f$ and $g$) each repeated consecutively in two homogeneous blocks gradually mixing positions to become an alternation between of the two symbols. DL distance is shown to have a wider range of values, while JW has more nuance registers more nuance on prefixes and positions. 

Our metrics for the string forecast accuracy are the normalized versions of the DL and JW distances computed using the Python library \verb|textdistance|\footnote{textdistance~4.2.0, \url{https://pypi.org/project/textdistance}}. The text similarity in this version gives a value of 1.0 for identical strings and a value of 0.0 for ``completely different'' strings.

\section{Results}

The following computational tests are performed on a Dell PowerEdge R740 Server with 1.5~TB RAM and two Intel Xeon Silver 4114 processors running at 2.2 GHz. The scripts were written and run in Python 3.7.3 using the libraries Pandas~1.2.3, NumPy~1.19.2, TensorFlow~2.4.1, and TextDistance~4.2.0. In order to reduce the number of parameter configurations to be studied, we have divided our tests into three parts. The first initial parameter study on medium-complexity seed strings will be used to fix the number of layers, decide on the stopping criterion for the training, and reduce the number of learning rates considered. The other two tests explore the remaining parameters with seed strings of low and high complexity, respectively. 

\subsection{Initial parameter test with medium complexity seed strings}
We start with an initial parameter study to set the basis for the following in-depth tests. For this test, 12 seed strings were generated using 2, 5, 10, and 20 symbols; with LZW complexities of 20, 35, and 50. Each of these seed strings was repeated to produce strings of at least 500 characters long. The trailing 100 characters of each of these strings are used as the validation data, while the other leading characters are used for the training. For each training string, an RNN is trained with various stopping criteria, learning rates, the number of layers, and units per layer. Each configuration is trained five times to reduce the effect of the random weight initialization. 

The Adam optimizer \citep{Kingma2015AdamAM} is used, motivated by the results of \cite{Ruder2017} who showed that adaptive learning-rate methods, and in particular Adam, yields the best results for sparse data such as one-hot encoded sequences. The learning rates are varied between $\{0.001, 0.01, 0.1\}$. 
The maximal number of training epochs is set to 999. Two stopping criteria are evaluated: (i) stop the training when the accuracy reaches a value larger or equal to 0.99, and (ii) stop when the loss function, in this case, categorical cross-entropy, reaches a value less or equal to 0.1. While the loss function is well known, it is worth mentioning that the aforementioned accuracy is calculated by computing the frequency in which the predicted values match the real $y$ values and dividing it by the total predictions, in this case, the total elements of $y$. 

\begin{figure}
	\centering
	\includegraphics[width=1.0\textwidth]{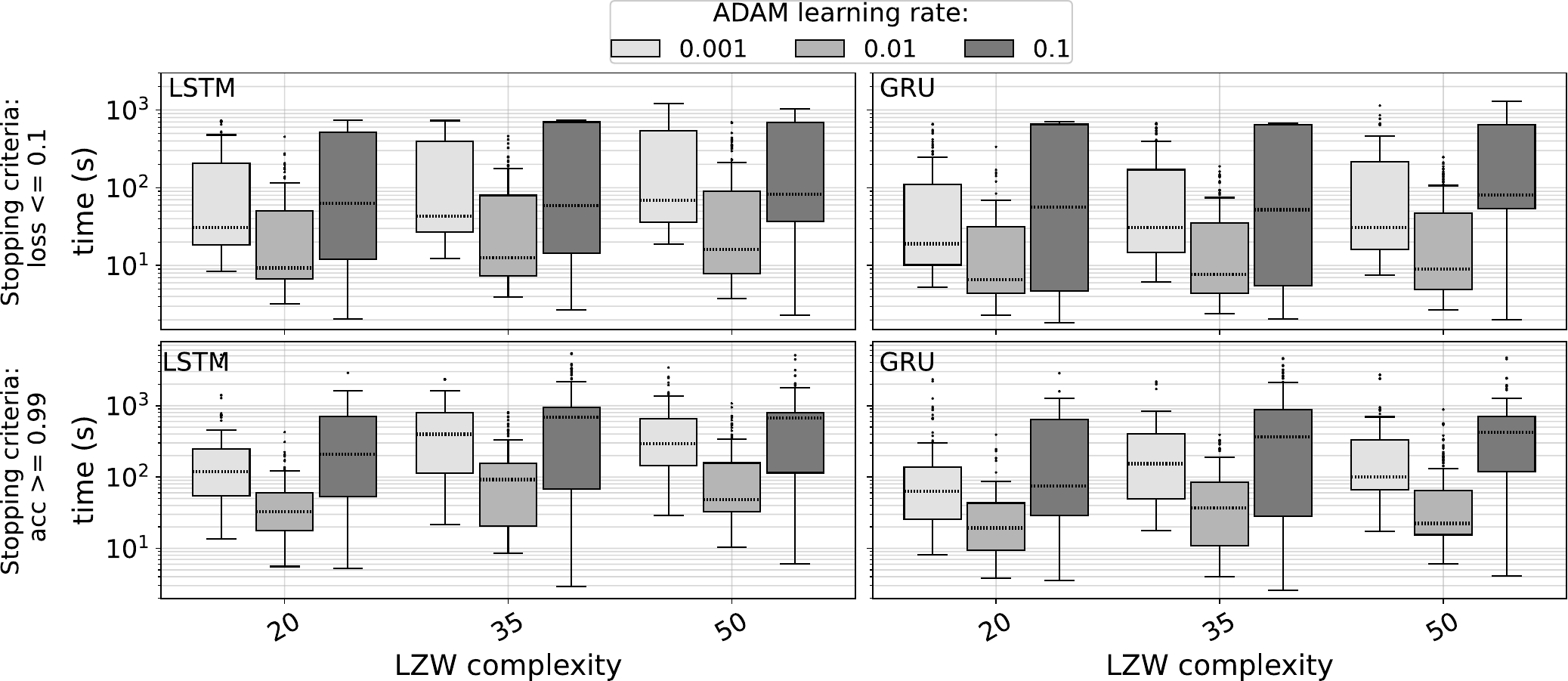}
	\caption[Total training time]{Total time needed for training LSTM and GRU networks on strings of different LZW complexities and with different stopping criteria and learning rates. The dotted line shows the median, the box represents the interquartile range (IQR, the middle 50\%), the whiskers have a length of 1.5$\cdot$IQR. All points outside the whiskers are considered outliers and are plotted individually. Note the logarithmic scale of the $y$-axis. A learning rate of 0.01 appears most suitable irrespective of the stopping criterion.}
	\label{fig:pre00}
\end{figure}

After the training is completed, a forecast of 100 characters is produced and its text similarity to the validation string is measured. For both stopping criteria, we found that a learning rate of 0.01 led to the smallest training times for all string complexities considered. This is summarized visually in Figure~\ref{fig:pre00}. 

\begin{figure}
	\centering
	\includegraphics[width=1.0\textwidth]{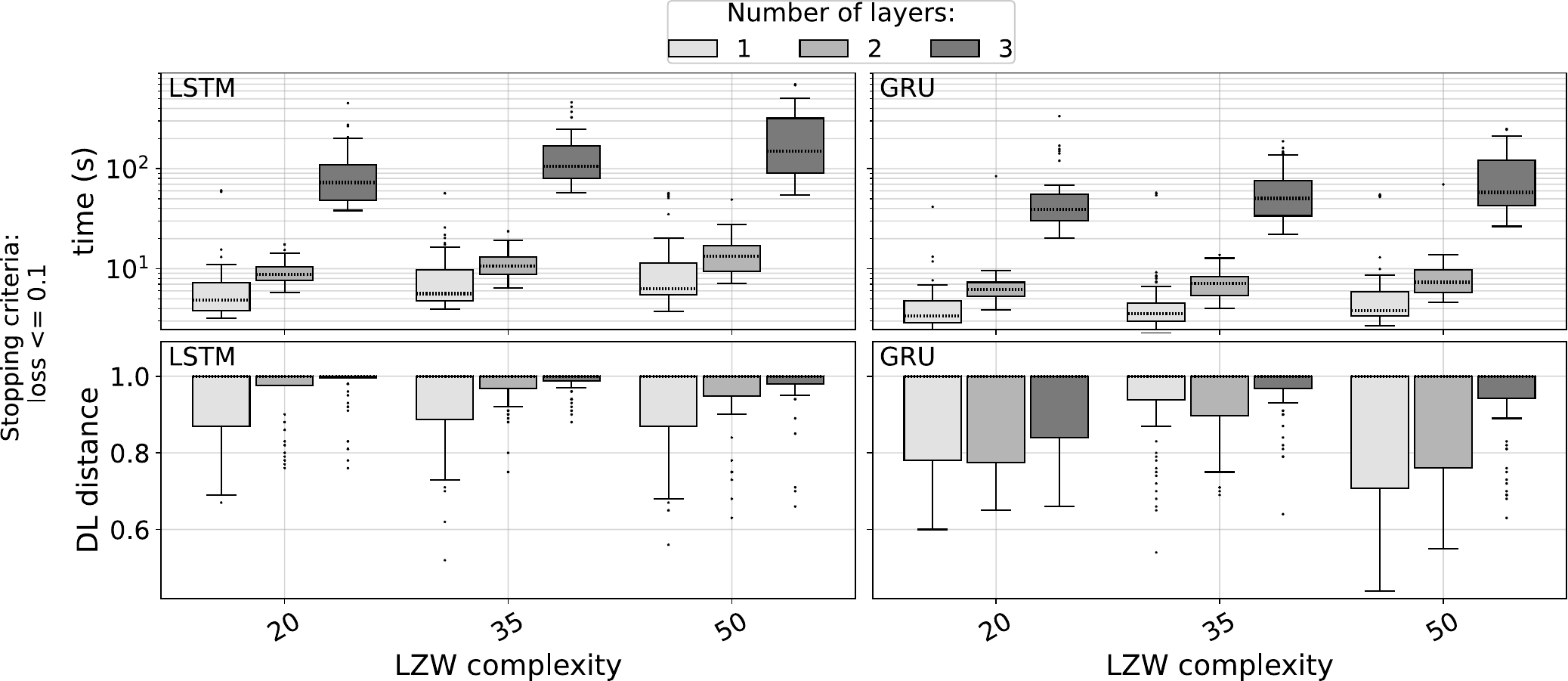}
	\caption[Dependency on the number of RNN layers]{Studying the dependency on the number of RNN layers, always using the same loss-based stopping criterion and a learning rate of 0.01. Note the logarithmic $y$-axes on the top. The addition of layers slightly increases accuracy in both DL and JW text similarities (here only DL is shown for simplicity) but the significant increase in training time makes it hard to justify the depth increase.}
	\label{fig:pre01}
\end{figure}

We next explore the string memorization capability of the networks dependent on the number of layers. We train LSTM and GRU networks with $\ell \in \{1,2,3\}$ layers and each layer having $u$ units, where $u$ is chosen such that $\ell u$ is closest to $\{50,100,200\}$ (i.e., the total number of units is approximately constant as $\ell$ varies). The quality of the forecasts measured using DL distances is averaged over all networks with the same number of layers and the whole 12 seed strings. In all cases, the loss-based stopping criterion is used, and the learning rate is~0.01. The results are shown in Figure~\ref{fig:pre01}. 
The most successful network configuration, in terms of small DL distance and training time, is a single hidden layer network (the results look similar for the JW distance). Although there is a slight improvement in forecast accuracy with each added hidden layer, the observed increase in training time does not seem to justify their addition. 

In summary, this initial parameter test trained 3,239 RNNs for the 12 different seed strings, over 5~runs to prevent outlier bias, and with the variety of parameters discussed above. The main finding is that the learning rate and the number of hidden units are among the most influential hyper-parameters for the effectiveness of the considered RNNs. This is consistent with findings in \cite{Greff2017}. In what follows, we will use single-layer RNNs with a reduced range of considered learning rates and perform larger studies with strings of lower and higher LZW complexities, respectively.

\subsection{Test with seed strings of low complexity}

In this low LZW complexity exploration, nearly 3,600 RNNs are trained for a total of 37 different seed strings. These seed strings are now repeated to produce sequences of a minimum length of 1,100 characters. Again, the trailing 100 characters are used as validation strings, with the remaining leading-strings used for the training and testing. The seed strings have LZW complexities ranging between 2 and 12 and are composed of numerous distinct symbols ranging between 2 and~6. A single hidden layer is used for both the LSTM and GRU networks. The number of units within the hidden layer is varied between 25 and 250, in ten geometrically-spaced steps. The Adam optimizer is used with learning rates of 0.001 and 0.01 and the aforementioned loss-based stopping criterion. As before, all configurations are run 5 times and averaged. 

\begin{figure}
	\centering
	\includegraphics[width=1.0\textwidth]{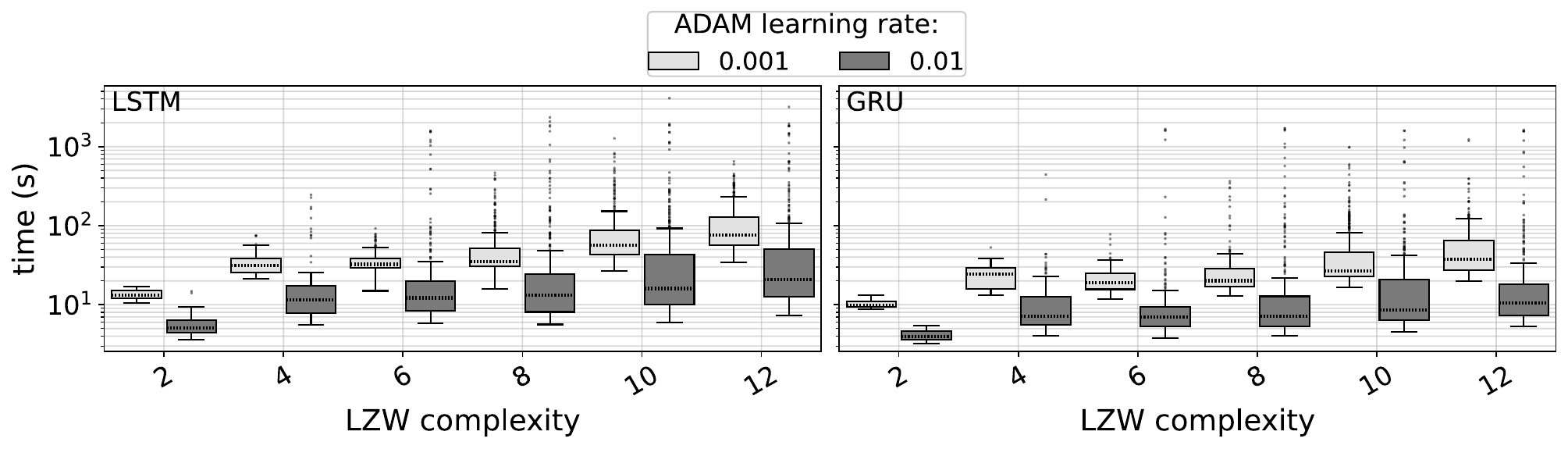}
	\caption[Training time on fitting low complexity seed strings]{Training time when fitting low complexity seed strings. On average, GRU requires about half the training time compared to LSTM.}
	\label{fig:LSTMvsGRU00}
\end{figure}

\begin{figure}
	\centering
	\includegraphics[width=1.0\textwidth]{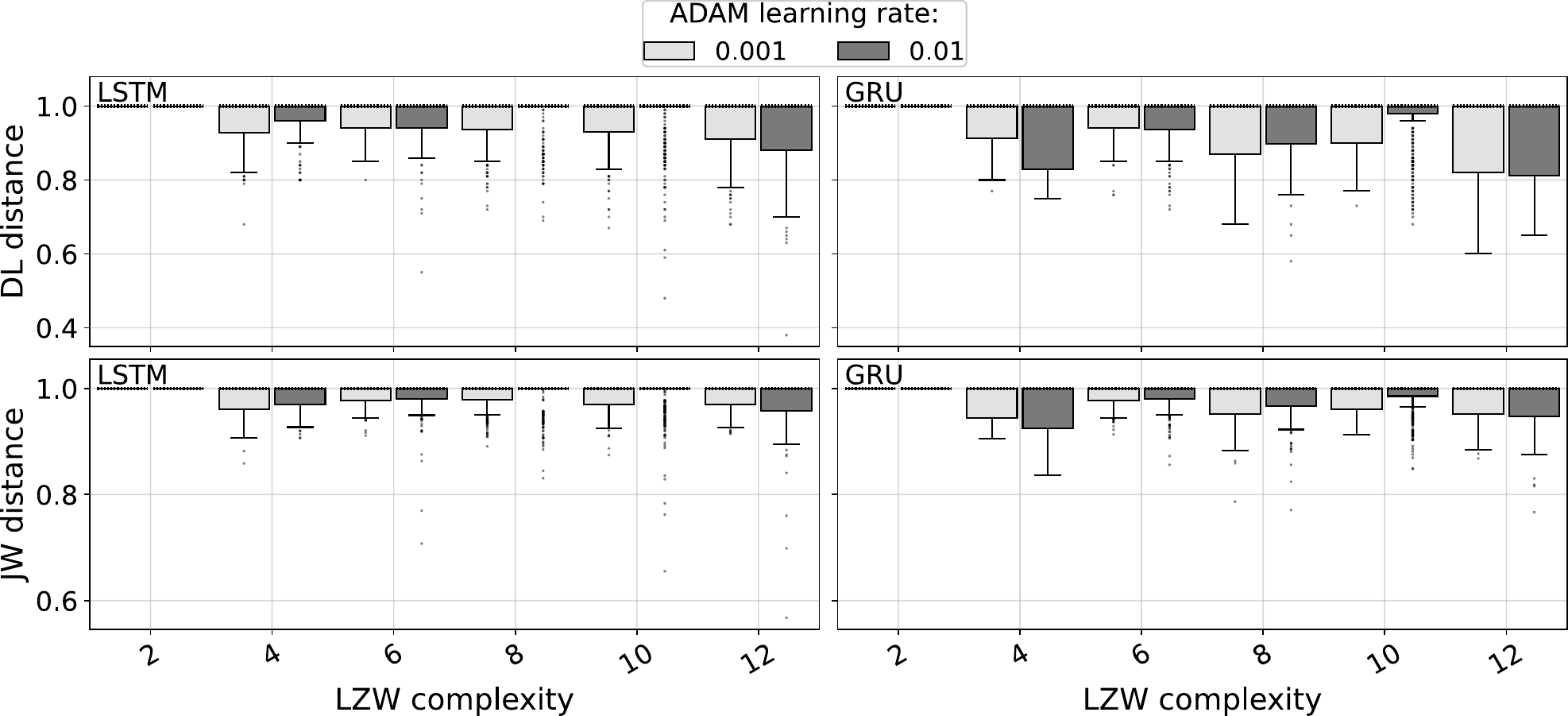}
	\caption[Performance on low complexity seed strings]{LSTM and GRU achieve similar forecast performance for low complexity seed strings, with LSTM sightly better but requiring more training time.}
	\label{fig:LSTMvsGRU01}
\end{figure}

A visual summary of the results is given in Figure~\ref{fig:LSTMvsGRU00}. The median training time for all tests with LSTM networks is 37.19 seconds with an interquartile range (IQR) between 15.64 to 75.79 seconds, and for GRU 19.72 seconds with an IQR between 8.48 to 31.70 seconds. We generally find that GRUs are trained faster than LSTM networks to achieve the same loss function value with the same optimizer overall considered learning rates and network complexities, not only in median values but also with less dispersion in general.
The forecast accuracy with LSTMs and GRUs is comparable. The median distance of both JW and DL metrics for both types of RNNs was 1.0, this is the same value for the third quartile ($Q_3$) however some differences are appreciated in the first quartile ($Q_1$). The values for LSTM are 0.93 and 0.97 for DL and JW distance respectively, whereas, for GRU, they are 0.88 for DL and 0.96 for JW. This tells us that despite longer training times LSTM seems to have a small advantage on accuracy. This information is presented visually in Figure~\ref{fig:LSTMvsGRU01}. One must remember that these text similarities are not Euclidean and small differences for JW usually correspond to more contrasting strings than DL.

\subsection{Test with seed strings of high complexity}

Our final study uses a total of 300 seed strings with 10, 33, or 52 symbols and LZW complexities ranging between 1,000 to 1,850 (168 linearly spaced steps between these bounds). The seed strings are all at most 2,400 characters long and then repeated to produce string sequences of 5,000, 7,500, and 10,000 characters, respectively. A total number of 4,500 RNNs is trained for this test.
We experienced some stagnation in the training of GRUs which was easily fixed by changing the learning rate from 0.01 to 0.0035, while for LSTM the learning rate of 0.01 was kept. The stopping criterion, number of units, and number of layers are fixed to loss-based, 100 and 1, respectively.

\begin{figure}
	\centering
	\includegraphics[width=.99\textwidth]{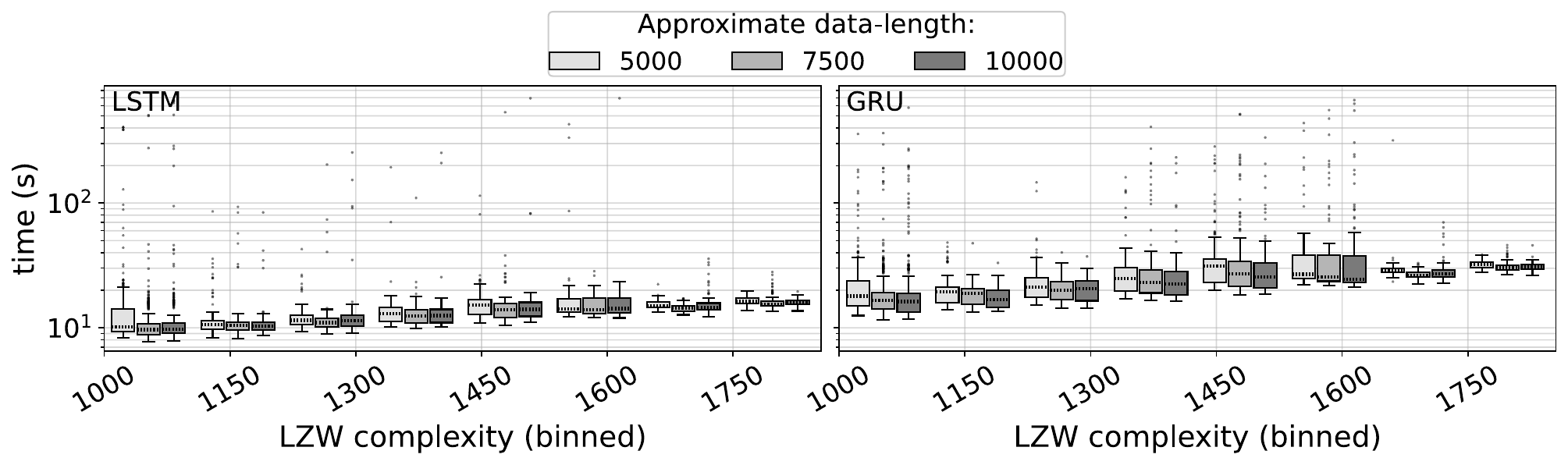}
	\caption[Performance on high complexity seed strings I]{Results for high complexity seed strings. Now LSTMs are faster to train than GRUs for a similar forecast performance.}
	\label{fig:LSTMvsGRU02}
\end{figure}
\begin{figure}
	\centering
	\includegraphics[width=.99\textwidth]{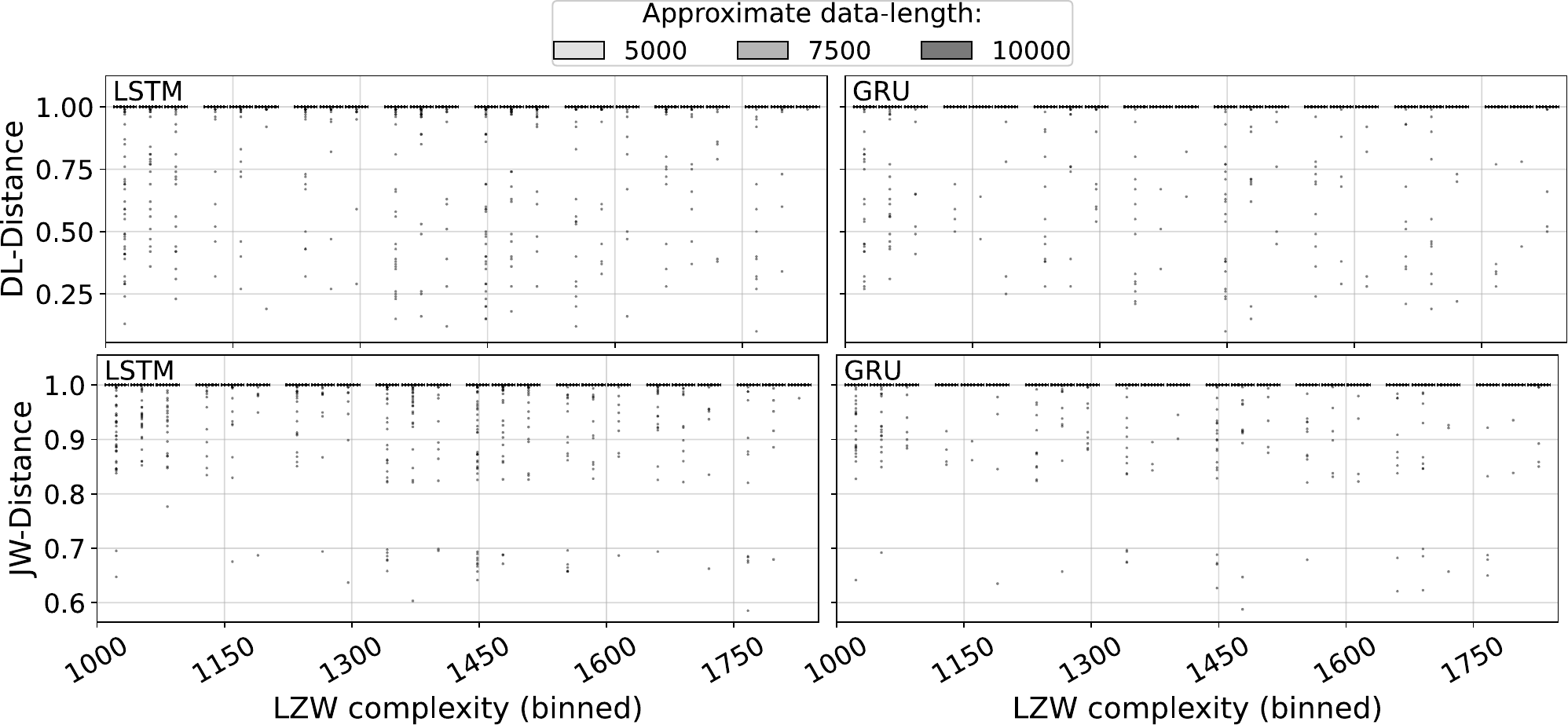}
	\caption[Performance on high complexity seed strings II]{Results for high complexity seed strings. LSTM and GRU achieve similar forecast accuracy in all cases (but LSTMs are faster to train; see Figure~\ref{fig:LSTMvsGRU02}).}
	\label{fig:LSTMvsGRU03}
\end{figure}

We find that LSTMs are better suited than GRUs for high complexity strings: the median training time is 12.53 seconds for LSTMs and almost double, namely 22.84 seconds, for GRUs, with an IQR between 10.59 and 15.07 and between 18.07 and 29.57 for LSTM and GRU, respectively; see Figure~\ref{fig:LSTMvsGRU02}. To simplify the plots, all 168 complexities were combined into 8 bins. Note that the data dispersion decreases drastically after the binned complexity of 1,600, which is caused by the larger number of seed strings using 52 symbols. The median and IQR values for both types of RNNs were found to be 1.0; see Figure~\ref{fig:LSTMvsGRU03}.

\section{Discussion}
We have used string sequences of quantifiable complexity to gain insights into hyper-parameter choices for two of the most common RNN variants. In terms of the string complexity, we found that the learning rate is a crucial parameter for efficient training and that an increase in RNN depth leads to a significant increase in training time but not necessarily forecast accuracy. Generally, GRUs outperformed LSTMs for low-complexity strings while LSTMs performed better on high-complexity strings. The latter finding is consistent with experiences in language modeling (typically involving very complex strings), where LSTM was also found to perform better than GRUs since it is better at capturing long-term dependencies \citep{irie2016lstm}.

\begin{figure}[ht]
	\centering
	\includegraphics[width=0.99\textwidth]{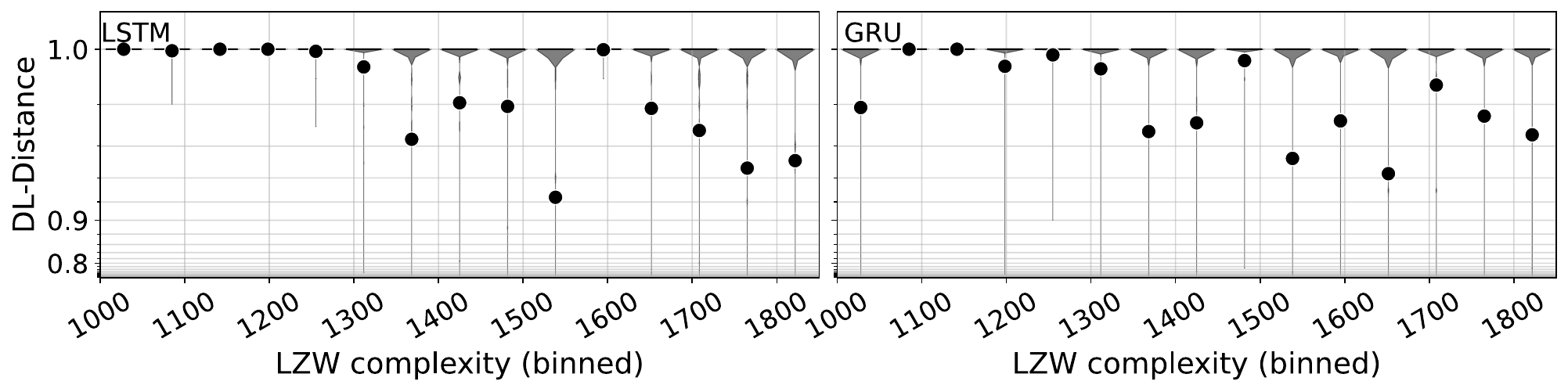}
	\caption[Saturation of learning capacity]{Saturation of learning capacity as the string complexity increases. Only seed strings of 52 symbols were considered in this test. The gray-shaded areas represent  KDEs with bandwidth 0.01, the markers represent the mean. The decreasing trend was observed for both LSTM and GRU with optimized hyper-parameters, trained for a maximum of 999 epochs. Note the exponential scale of the x-axis to emphasize the decrease in mean text similarity,  suggesting a degradation of the forecasting quality.}
	\label{fig:saturation00}
\end{figure}

In all our tests the networks have been able to learn all sequences with relatively high accuracy. This need not be the case, however: if the complexity of a string becomes very high, the network's learning capability might be restricted. This is manifested by an observed decrease in the mean values of text similarity and an increase in outlier scattering. A demonstration of this is shown in~\figurename~\ref{fig:saturation00}.

\bibliographystyle{spbasic.bst} 
\bibliography{ref.bib}

\begin{thebibliography}{31}
\providecommand{\natexlab}[1]{#1}
\providecommand{\url}[1]{{#1}}
\providecommand{\urlprefix}{URL }
\expandafter\ifx\csname urlstyle\endcsname\relax
  \providecommand{\doi}[1]{DOI~\discretionary{}{}{}#1}\else
  \providecommand{\doi}{DOI~\discretionary{}{}{}\begingroup
  \urlstyle{rm}\Url}\fi
\providecommand{\eprint}[2][]{\url{#2}}

\bibitem[{Bandara et~al(2020)Bandara, Bergmeir, and Smyl}]{Bandara2020}
Bandara K, Bergmeir C, Smyl S (2020) Forecasting across time series databases
  using recurrent neural networks on groups of similar series: A clustering
  approach. Expert Systems with Applications 140:112,896

\bibitem[{Boytsov(2011)}]{Boystov2011}
Boytsov L (2011) Indexing methods for approximate dictionary searching:
  Comparative analysis. Journal of Experimental Algorithmics 16:1.10--1.91

\bibitem[{Cahuantzi et~al(2021)Cahuantzi, Chen, and
  G\"{u}ttel}]{Cahuantzi_slearn_2021}
Cahuantzi R, Chen X, G\"{u}ttel S (2021) {slearn}.
  \urlprefix\url{https://github.com/nla-group/slearn}

\bibitem[{Cho et~al(2014)Cho, van Merri{\"e}nboer, Gulcehre, Bahdanau,
  Bougares, Schwenk, and Bengio}]{cho2014}
Cho K, van Merri{\"e}nboer B, Gulcehre C, Bahdanau D, Bougares F, Schwenk H,
  Bengio Y (2014) Learning phrase representations using {RNN}
  encoder{--}decoder for statistical machine translation. In: Proceedings of
  the 2014 Conference on Empirical Methods in Natural Language Processing, pp
  1724--1734

\bibitem[{Elsworth and Güttel(2020)}]{Elsworth2020lstm}
Elsworth S, Güttel S (2020) Time series forecasting using {LSTM} networks: {A}
  symbolic approach. arXiv 2003.05672

\bibitem[{Greff et~al(2017)Greff, Srivastava, Koutnik, Steunebrink, and
  Schmidhuber}]{Greff2017}
Greff K, Srivastava RK, Koutnik J, Steunebrink BR, Schmidhuber J (2017) {LSTM}:
  A search space odyssey. Transactions on Neural Networks and Learning Systems
  28:2222--2232

\bibitem[{Hochreiter and Schmidhuber(1997)}]{Hochreiter1997}
Hochreiter S, Schmidhuber J (1997) Long short-term memory. Neural Computation
  9:1735--1780

\bibitem[{Irie et~al(2016)Irie, T{\"u}ske, Alkhouli, Schl{\"u}ter, and
  Ney}]{irie2016lstm}
Irie K, T{\"u}ske Z, Alkhouli T, Schl{\"u}ter R, Ney H (2016) {LSTM}, {GRU},
  highway and a bit of attention: An empirical overview for language modeling
  in speech recognition. In: Interspeech, pp 3519--3523

\bibitem[{Jozefowicz et~al(2015)Jozefowicz, Zaremba, and
  Sutskever}]{pmlr-v37-jozefowicz15}
Jozefowicz R, Zaremba W, Sutskever I (2015) An empirical exploration of
  recurrent network architectures. In: Proceedings of the 32nd International
  Conference on Machine Learning, PMLR, Proceedings of Machine Learning
  Research, vol~37, pp 2342--2350

\bibitem[{Kaspar and Schuster(1987)}]{Kaspar1987}
Kaspar F, Schuster HG (1987) Easily calculable measure for the complexity of
  spatiotemporal patterns. Physical Review A 36:842--848

\bibitem[{Kim et~al(2021)Kim, Nam, Kim, and Jung}]{Kim_Nam_Kim_Jung_2021}
Kim S, Nam H, Kim J, Jung K (2021) Neural sequence-to-grid module for learning
  symbolic rules. Proceedings of the AAAI Conference on Artificial Intelligence
  35(9):8163--8171

\bibitem[{Kingma and Ba(2015)}]{Kingma2015AdamAM}
Kingma DP, Ba J (2015) Adam: A method for stochastic optimization. In:
  International Conference on Learning Representations

\bibitem[{Kolmogorov(1963)}]{Kolmogorov1963}
Kolmogorov AN (1963) On tables of random numbers. Sankhyā: The Indian Journal
  of Statistics, Series A 25:369--376

\bibitem[{Lample and Charton(2020)}]{Lample2020Deep}
Lample G, Charton F (2020) Deep learning for symbolic mathematics. In:
  International Conference on Learning Representations

\bibitem[{Lin et~al(2003)Lin, Keogh, Lonardi, and Chiu}]{Wang2005}
Lin J, Keogh E, Lonardi S, Chiu B (2003) A symbolic representation of time
  series, with implications for streaming algorithms. In: SIGMOD Workshop on
  Research Issues in Data Mining and Knowledge Discovery, ACM, pp 2--11

\bibitem[{Lin et~al(2017)Lin, Guo, and Aberer}]{10.5555/3172077.3172204}
Lin T, Guo T, Aberer K (2017) Hybrid neural networks for learning the trend in
  time series. In: International Joint Conference on Artificial Intelligence,
  pp 2273--2279

\bibitem[{Maas et~al(2012)Maas, Le, O’Neil, Vinyals, Nguyen, and Ng}]{45168}
Maas A, Le QV, O’Neil TM, Vinyals O, Nguyen P, Ng AY (2012) Recurrent neural
  networks for noise reduction in robust {ASR}. In: INTERSPEECH

\bibitem[{Mao et~al(2015)Mao, Xu, Yang, Wang, Huang, and Yuille}]{mao2014deep}
Mao J, Xu W, Yang Y, Wang J, Huang Z, Yuille A (2015) Deep captioning with
  multimodal recurrent neural networks (m-rnn). International Conference on
  Learning Representations

\bibitem[{Montero-Manso and Hyndman(2020)}]{Monteromanso2021}
Montero-Manso P, Hyndman RJ (2020) Principles and algorithms for forecasting
  groups of time series: Locality and globality. Tech. rep., Monash University,
  Department of Econometrics and Business Statistics

\bibitem[{Oreshkin et~al(2020)Oreshkin, Carpov, Chapados, and
  Bengio}]{Oreshkin2020NBEATSNB}
Oreshkin BN, Carpov D, Chapados N, Bengio Y (2020) {N-BEATS}: Neural basis
  expansion analysis for interpretable time series forecasting. In:
  International Conference on Learning Representations

\bibitem[{Rabanser et~al(2020)Rabanser, Januschowski, Flunkert, Salinas, and
  Gasthaus}]{rabanser2020}
Rabanser S, Januschowski T, Flunkert V, Salinas D, Gasthaus J (2020) The
  effectiveness of discretization in forecasting: An empirical study on neural
  time series models. arXiv 2005.10111

\bibitem[{Ruder(2016)}]{Ruder2017}
Ruder S (2016) An overview of gradient descent optimization algorithms. arXiv
  1609.04747

\bibitem[{Salinas et~al(2020)Salinas, Flunkert, Gasthaus, and
  Januschowski}]{SALINAS20201181}
Salinas D, Flunkert V, Gasthaus J, Januschowski T (2020) {DeepAR}:
  Probabilistic forecasting with autoregressive recurrent networks.
  International Journal of Forecasting 36(3):1181--1191

\bibitem[{Smyl(2020)}]{smyl2020hybrid}
Smyl S (2020) A hybrid method of exponential smoothing and recurrent neural
  networks for time series forecasting. International Journal of Forecasting
  36:75--85

\bibitem[{Venugopalan et~al(2015)Venugopalan, Rohrbach, Donahue, Mooney,
  Darrell, and Saenko}]{7410872}
Venugopalan S, Rohrbach M, Donahue J, Mooney R, Darrell T, Saenko K (2015)
  Sequence to sequence -- video to text. In: International Conference on
  Computer Vision, IEEE, pp 4534--4542

\bibitem[{Welch(1984)}]{Welch1984}
Welch T (1984) A technique for high-performance data compression. Computer
  17:8--19

\bibitem[{Winkler(2006)}]{Winkler2006}
Winkler WE (2006) Overview of record linkage and current research directions.
  Tech. rep., Bureau of the Census

\bibitem[{Yamak et~al(2019)Yamak, Yujian, and Gadosey}]{yamak2019comparison}
Yamak PT, Yujian L, Gadosey PK (2019) A comparison between {ARIMA}, {LSTM}, and
  {GRU} for time series forecasting. In: International Conference on
  Algorithms, Computing and Artificial Intelligence, ACM, pp 49--55

\bibitem[{Zaremba et~al(2014)Zaremba, Kurach, and Fergus}]{NIPS2014_08419be8}
Zaremba W, Kurach K, Fergus R (2014) Learning to discover efficient
  mathematical identities. In: Ghahramani Z, Welling M, Cortes C, Lawrence N,
  Weinberger K (eds) Advances in Neural Information Processing Systems, Curran
  Associates, Inc., vol~27

\bibitem[{Zenil(2020)}]{Zenil2020}
Zenil H (2020) A review of methods for estimating algorithmic complexity:
  Options, challenges, and new directions. Entropy 22:1--28

\bibitem[{Zhang et~al(2017)Zhang, Bahrampour, Ramakrishnan, Schott, and
  Shah}]{10.1109/ICASSP.2017.7953302}
Zhang S, Bahrampour S, Ramakrishnan N, Schott L, Shah M (2017) Deep learning on
  symbolic representations for large-scale heterogeneous time-series event
  prediction. In: International Conference on Acoustics, Speech and Signal
  Processing, IEEE, pp 5970--5974

\end{thebibliography}

\end{document}